\documentclass{article}

\PassOptionsToPackage{numbers}{natbib}

\usepackage[preprint]{neurips_2023}

\usepackage[utf8]{inputenc} %
\usepackage[T1]{fontenc}    %
\usepackage[colorlinks]{hyperref}       %
\usepackage{url}            %
\usepackage{nicefrac}       %
\usepackage{microtype}      %
\usepackage{graphicx}
\usepackage{amsmath}
\usepackage{amssymb}
\usepackage{booktabs}
\usepackage{microtype}
\usepackage{xspace}
\usepackage{amsmath,amsfonts,amssymb,amsthm}
\usepackage[dvipsnames]{xcolor}
\usepackage{diagbox}
\usepackage{pifont}
\usepackage{svg}
\usepackage{amsmath}
\usepackage{enumitem}
\usepackage{physics}
\usepackage{mathtools}
\usepackage{algorithm}
\usepackage{algorithmic}
\usepackage{gensymb}
\usepackage{wrapfig}
\usepackage{subcaption}
\usepackage{multirow}
\usepackage{footmisc}
\usepackage{empheq}
\usepackage{cases}
\usepackage{adjustbox}
\usepackage{array}
\usepackage{tabu}
\usepackage{flushend}
\usepackage{multirow}
\usepackage{bm}
\usepackage{color, colortbl}
\usepackage[numbers]{natbib}

\title{

TOAST: Transfer Learning via Attention Steering
}

\author{%
  Baifeng Shi \\
  UC Berkeley \\
  \And
  Siyu Gai \\
  UC Berkeley \\
  \And
  Trevor Darrell \\
  UC Berkeley \\
  \And
  Xin Wang \\
  Microsoft Research \\
}

\begin{document}

\makeatletter
\DeclareRobustCommand\onedot{\futurelet\@let@token\@onedot}
\def\@onedot{\ifx\@let@token.\else.\null\fi\xspace}

\def\eg{\emph{e.g.}\xspace} \def\Eg{\emph{E.g}\onedot}
\def\ie{\emph{i.e.}\xspace} \def\Ie{\emph{I.e}\onedot}
\def\vs{\emph{vs.}\xspace}
\def\cf{\emph{c.f}\onedot} \def\Cf{\emph{C.f}\onedot}
\def\etc{\emph{etc}\onedot} \def\vs{\emph{vs}\onedot}
\def\wrt{w.r.t\onedot} \def\dof{d.o.f\onedot}
\def\etal{\emph{et al}\onedot}
\def\viz{\emph{viz}\onedot}
\makeatother

\newcommand{\cmark}{\ding{51}}%
\newcommand{\xmark}{\ding{55}}%

\newcommand{\bbR}{\mathbb{R}}
\newcommand{\bfx}{\mathbf{x}}
\newcommand{\bfh}{\mathbf{h}}
\newcommand{\bfz}{\mathbf{z}}
\newcommand{\bfZ}{\mathbf{Z}}
\newcommand{\tbfz}{\widetilde{\mathbf{z}}}
\newcommand{\tbfZ}{\widetilde{\mathbf{Z}}}
\newcommand{\bfu}{\mathbf{u}}
\newcommand{\bfU}{\mathbf{U}}
\newcommand{\tbfu}{\widetilde{\mathbf{u}}}
\newcommand{\tbfU}{\widetilde{\mathbf{U}}}
\newcommand{\bfQ}{\mathbf{Q}}
\newcommand{\bfK}{\mathbf{K}}
\newcommand{\bfV}{\mathbf{V}}
\newcommand{\bfP}{\mathbf{P}}
\newcommand{\bfJ}{\mathbf{J}}
\newcommand{\bfW}{\mathbf{W}}
\newcommand{\bfX}{\mathbf{X}}

\newcommand{\method}{{TOAST}\xspace} 
\newcommand{\methodlite}{{TOAST-Lite}\xspace} 
\newcommand\minisection[1]{\vspace{1.3mm}\noindent \textbf{#1}}

\newcommand{\todo}[1]{{\color{purple}{\bf TODO: #1}}}

\definecolor{Gray}{gray}{0.9}

\setcounter{topnumber}{3}

\maketitle

\begin{abstract}

Transfer learning involves adapting a pre-trained model to novel downstream tasks. However, we observe that current transfer learning methods often fail to focus on task-relevant features. In this work, we explore refocusing model attention for transfer learning. We introduce \textbf{To}p-Down \textbf{A}ttention \textbf{St}eering (\textbf{\method}), a novel transfer learning algorithm that keeps the pre-trained backbone frozen, selects task-relevant features in the output, and feeds those features back to the model to steer the attention to the task-specific features. By refocusing the attention only, \method achieves state-of-the-art results on a number of transfer learning benchmarks, while having a small number of tunable parameters. Compared to fully fine-tuning, LoRA, and prompt tuning, \method substantially improves performance across a range of fine-grained visual classification datasets (\eg, $81.1\%$ $\rightarrow$ $86.2\%$ on FGVC). \method also outperforms the fully fine-tuned Alpaca and Vicuna models on instruction-following language generation. Code is available at \url{https://github.com/bfshi/TOAST}.

\end{abstract}

\section{Introduction}

The prevailing approach for addressing novel tasks with deep learning is leveraging a pre-trained model and transferring it to the specific downstream task~\cite{bommasani2021opportunities}. Common approaches for transfer learning involve updating parts or all of the parameters in the model (\eg, fine-tuning, LoRA~\cite{hu2021lora}) or adding task-specific augmentation to the input (\eg, prompt tuning~\cite{lester2021power}, VPT~\cite{jia2022visual}) in order to adjust model features for the downstream task. 

\begin{figure}[t]
  \begin{center}
      \includegraphics[width=0.7\linewidth]{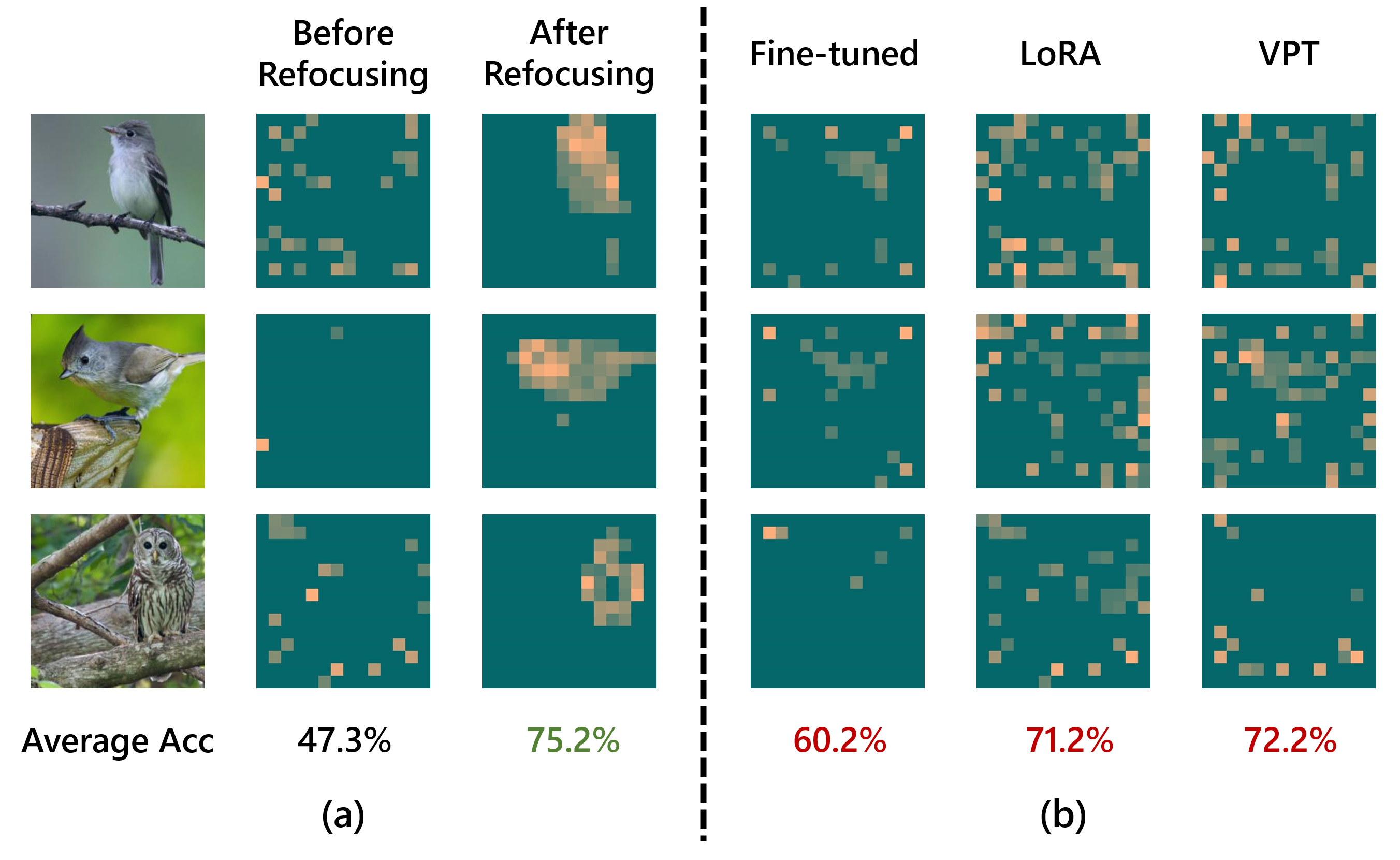}
  \end{center}
  \vspace{-1em}
  \caption{We take an ImageNet pre-trained ViT and transfer it to downstream bird classification using different transfer learning algorithms. Here we visualize the attention maps of these models. Each attention map is averaged across different heads in the last layer of ViT. \textbf{(a)} Our method, \method, is able to refocus the attention of a pre-trained backbone onto task-specific features, improving the downstream performance by a large margin. \textbf{(b)} Previous transfer learning methods such as fine-tuning, LoRA, and VPT fail to focus on task-relevant objects, achieving suboptimal performance.}
  \label{fig:intro}
  \vspace{-1em}
\end{figure}

However, we empirically find that previous transfer learning methods usually fail to focus the model's attention on task-relevant signals. For example, in Figure~\ref{fig:intro}(b) we visualize the attention map of a ViT model pre-trained on ImageNet and transferred to downstream bird classification via fine-tuning, LoRA, or VPT. Such attention maps are often extremely noisy and fail to focus on task-specific objects. This encourages us to rethink the role of attention in transfer learning and if we can boost performance by refocusing the model's attention on task-related signals.

In this work, we show that \emph{refocusing attention is key to transfer learning}. We introduce \textbf{To}p-Down \textbf{A}ttention \textbf{St}eering (\textbf{\method}), a novel transfer learning approach that learns new tasks by redirecting the attention to task-relevant features. This is achieved through a top-down attention module~\cite{shi2023top} which allows a model to adjust its attention in a task-adaptive way. The top-down attention module takes the output features from the backbone, selects the features that are relevant to the task, and then feeds them back into each self-attention layer in the backbone. These top-down signals will enhance the task-relevant features in each layer, and the feedforward backbone runs again with the enhanced feature, achieving stronger attention on the task-relevant signals. When transferring to different downstream tasks, \method simply freezes the pre-trained backbone and tunes the top-down attention module to steer the attention to task-specific signals (Figure~\ref{fig:intro}(a)).

Remarkably, by simply refocusing attention, \method achieves state-of-the-art results on various transfer learning benchmarks. Compared to fully fine-tuning, LoRA, and VPT, \method significantly improves the performances on FGVC fine-grained classification (\eg, $5\%$ improvement over fully fine-tuning on average accuracy), and obtains the best performance on 11 out of 18 tasks on VTAB benchmark~\cite{zhai2019large}. Beyond visual recognition, \method can adapt large language models such as LLaMA-7B~\cite{touvron2023llama} for instruction-following language generation, resulting in more detailed and informed answers and outperforming fully fine-tuned Alpaca~\cite{alpaca}. We also explore \methodlite, a more parameter-efficient version of \method that tunes a similar number of parameters as LoRA while reaching higher performances. These observations strengthen our idea that refocusing attention is key to transfer learning and sheds light on future exploration in the field.

\section{Related Work}

\minisection{Transfer learning} refers to adapting a pre-trained model to a downstream task, which has become the paradigm of tackling unseen tasks for both vision~\cite{zhuang2020comprehensive} and language~\cite{devlin2018bert,radford2018improving}. Normal approaches for transfer learning involve tuning all the parameters (\ie, fully fine-tuning) or part of the parameters (\eg, only tuning the last few layers~\cite{yosinski2014transferable} or the bias terms~\cite{cai2020tinytl} of the network). Recent progress on large foundation models~\cite{bommasani2021opportunities,dehghani2023scaling,touvron2023llama} has promoted the exploration of Parameter-Efficient Fine-Tuning (PEFT) which is able to adapt the model by tuning only a small number of parameters (usually less than 1\% of all the parameters) and thus is more suitable for large models with billions of parameters. Common strategies include freezing the pre-trained backbone and adding additional tunable parameters (\eg, Adapter~\cite{houlsby2019parameter}, LoRA~\cite{hu2021lora}) or task-specific input augmentation (\eg, prefix tuning~\cite{liu2021p}, prompt tuning~\cite{lester2021power,jia2022visual}). However, fully fine-tuning usually obtains the highest empirical performance compared to other methods~\cite{ding2022delta}. In this work, we break the trade-off of downstream performance and parameter efficiency and show that our proposed method is able to outperform fine-tuning while having fewer tunable parameters.

\minisection{Top-down attention and its relation to transfer learning}. Top-down attention, one of the hallmarks of the human visual system, is the ability to selectively focus one's attention on the input signals that are relative to the current task or goal~\cite{carrasco2011visual,li2014understanding}. Top-down attention has been applied to different computer vision tasks such as object detection~\cite{oliva2003top}, image captioning~\cite{xu2015show}, and visual question answering~\cite{anderson2018bottom,xu2016ask}. Recent work~\cite{shi2023top} has designed a top-down attention module for Transformer, which we adopt in this work. On the other hand, previous studies on human perceptual learning have indicated a close relationship between top-down attention and how humans adapt to unseen tasks. Specifically, top-down attention facilitates learning new tasks by extracting task-relevant features while ignoring the distracting information~\cite{lavie1995perceptual,rees1997modulating}. Additionally, it is shown that only the task-relevant features are enhanced during adaptation, while the irrelevant features remain undistorted~\cite{ahissar1997task,schoups2001practising,tsushima2009roles}. This stands in contrast with transfer learning algorithms such as fully fine-tuning where all the pre-trained features are modified, indicating that the key to learning new tasks is adjusting the attention, not the pre-trained features.

\section{Top-Down Attention Steering}

\begin{figure}[t]
  \includegraphics[width=1\linewidth]{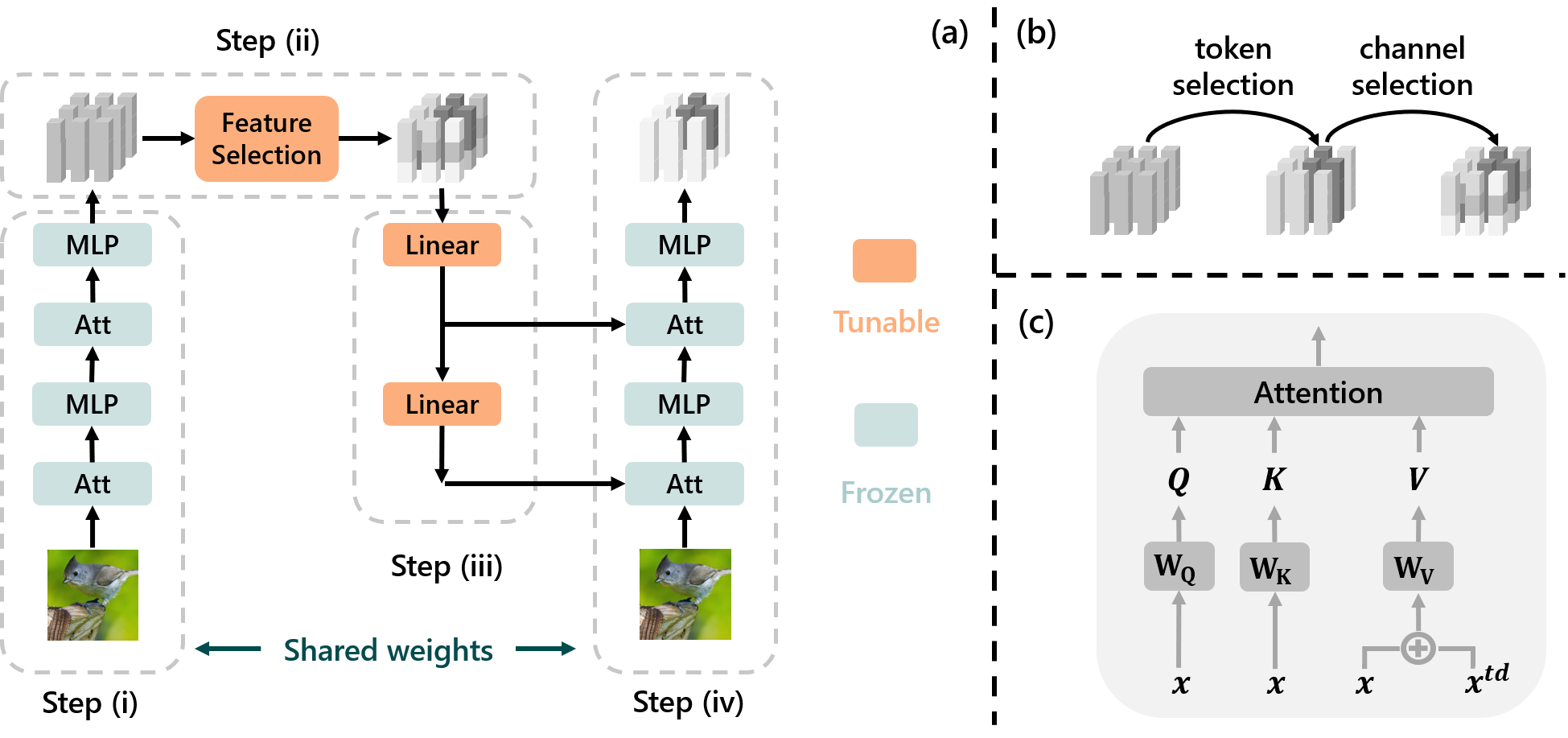}
  \vspace{-0.5em}
  \caption{\textbf{(a)} Overview of \method. In addition to the regular feedforward transformer which contains interleaving MLP and Attention blocks, we add a feature selection module and a feedback path consisting of linear layers. Inference has four steps: (i) the input goes through the feedforward transformer, (ii) the output tokens are softly reweighted by the feature selection module based on their relevance to the task, (iii) the reweighted tokens are sent back through the feedback path, and (iv) we run the feedforward pass again but with each attention layer receiving additional top-down inputs. During the transfer, we only tune the features selection module and the feedback path and keep the feedforward backbone frozen. \textbf{(b)} The feature selection module first selects the task-relevant tokens by reweighting the tokens based on their similarity to the task embedding, then selects the task-relevant channels by applying a task-specific linear transform on the channel dimension. \textbf{(c)} In the second feedforward pass, each self-attention layer receives an additional top-down input, which is added on the value matrix. }
  \label{fig:toat}
\end{figure}

We propose \textbf{To}p-Down \textbf{A}ttention \textbf{St}eering (\textbf{\method}), a novel transfer learning method that arms the pre-trained model with a top-down attention module and only tunes the top-down attention when transferring to downstream tasks. We first give a preliminary introduction to top-down attention (Section \ref{sec:method_arch}), then describe the detailed pipeline of \method (Section \ref{sec:method_toast}). Note that although \method is applicable to different model architectures such as transformers~\cite{dosovitskiy2020image,vaswani2017attention} and Convnets~\cite{liu2022convnet} as shown in Section~\ref{sec:exp_versatile}, we assume a transformer backbone in the following discussion.

\subsection{Preliminary: Transformer With Top-Down Attention}
\label{sec:method_arch}

Transformer model is usually bottom-up, \ie, its attention only depends on the input, and as a consequence, it normally highlights all the salient features in the input signal. As opposed to bottom-up attention, top-down attention endows the ability to adapt one's attention according to the high-level goal or task, \ie, it only focuses on the task-relevant features while ignoring the others~\cite{carrasco2011visual,li2014understanding}.

In this work, we follow the top-down attention design proposed in~\cite{shi2023top}, which is illustrated in Figure~\ref{fig:toat}(a). Specifically, for a regular transformer which is purely feedforward, we add a feature selection module and a feedback path for top-down attention. Inference of the network contains four steps: \textbf{(i)} pass the input through the feedforward path to get an initial output, \textbf{(ii)} select which features in the output is useful for the current task, \textbf{(iii)} the selected features are passed through the feedback path and sent back to each self-attention module, and \textbf{(iv)} run the feedforward pass again but with each self-attention receiving the top-down signal as additional input. In this way, the task-relevant information 
 is enhanced in each layer, achieving top-down attention.

Within the network, the feedforward path is a regular transformer, and the rest is described below:

\minisection{Feature selection (Step (ii))}. From the output of the feedforward backbone, this module selects the features that are useful for the task at hand. This includes the selection of both the tokens and the channels that are task-relevant. Figure~\ref{fig:toat}(b) illustrate the process. Specifically, denoting the output from the first feedforward pass by $(\bfz_i)_{i=1}^N$ where $\bfz_i \in \mathbb{R}^d$ is the $i$-th output token, the feature selection operates on each token and outputs $\tbfz_i = \bfP \cdot sim(\bfz_i, \xi) \cdot \bfz_i$, where $\xi \in \mathbb{R}^d$ and $\bfP \in \mathbb{R}^{d \times d}$ are task-specific parameters, and $sim(\cdot, \cdot)$ is cosine similarity clamped to $[0, 1]$. Here $\xi$ acts as a task embedding that encodes what kind of tokens are important for the task, and each token $\bfz_i$ is reweighted by its relevance (measured by cosine similarity) with the task embedding, simulating the token selection. Then the linear transform by $\bfP$ executes the channel selection for each token.

\minisection{Feedback path (Step (iii))}. After feature selection, the output tokens are sent back through the feedback path. The feedback path contains the same number of layers as the feedforward path, and each layer is a simple linear transform. The output from each layer goes through another linear transform and is sent into the self-attention module as the top-down input for the second feedforward.

\minisection{Self-attention with top-down input (Step (iv))}. In the second feedforward pass, each self-attention module receives an additional top-down input. As shown in Figure~\ref{fig:toat}(c), we simply add it to the value matrix while keeping the query and key untouched, \ie, $\bfQ, \bfK, \bfV = \bfW_Q \bfX, \bfW_K \bfX, \bfW_V (\bfX + \bfX^{td})$, where $\bfX$ is the regular bottom-up input to the self-attention module, and $\bfX^{td}$ is the top-down input. Then the regular self-attention on $\bfQ, \bfK, \bfV$ follows.

\subsection{Top-Down Attention Steering}
\label{sec:method_toast}

Given a pre-trained transformer, \method randomly initialize a top-down attention module and follows a two-stage pipeline: \textbf{(i)} pre-tuning the top-down attention on a general public dataset (\eg, ImageNet~\cite{deng2009imagenet} for vision or OpenWebText~\cite{Gokaslan2019OpenWeb} for language) to get a better initialization, and \textbf{(ii)} tuning the top-down attention on the downstream task. In both stages, we freeze the pre-trained backbone and only tune the top-down attention module (Figure~\ref{fig:toat}(a)).

\minisection{Pre-tuning stage}. Since the top-down attention module is randomly initialized, directly tuning it on downstream tasks might lead to suboptimal performance (see ablation in Section~\ref{sec:exp_ablation}). To this end, we propose to first pre-tune the top-down attention on a general public dataset such as ImageNet or OpenWebText to get a better initialization. During pre-tuning, except for the regular supervised or unsupervised loss, we also add the variational loss proposed in \cite{shi2023top}, which encourages the feedback path to reconstruct the input from the output, acting as a regularization on the feedback weights.

\minisection{Tuning stage}. When transferring to the downstream task, \method only fine-tunes the parameters in the top-down attention module. In this case, around $15\%$ of the parameters are updated. We notice that most of the tunable parameters are from the feedback layers, each of which contains a $d\times d$ matrix and is large when the feature dimension $d$ is high. To further promote parameter efficiency, we also propose \methodlite, which applies LoRA on the feedback layers. In this way, only less than $1\%$ of the parameters are tuned. We empirically show that \methodlite performs on par with \method on certain tasks while slightly worse on others (see Section~\ref{sec:exp_pe}).

\section{Experiments}
\label{sec:exp}

In this section, we first try to understand the attention refocusing process in \method by visualizing the attention maps (Section~\ref{sec:exp_understand}). Then we evaluate \method's performance on visual classification (Section~\ref{sec:exp_vision}), language generation (Section~\ref{sec:exp_language}), as well as its versatility when adapting to different tasks and model architectures (Section~\ref{sec:exp_versatile}). We also evaluate \methodlite, the parameter-efficient version of \method (Section~\ref{sec:exp_pe}). Finally, we conduct ablation studies on the designing choices of \method (Section~\ref{sec:exp_ablation}).

\minisection{Datasets}. We pre-tune the top-down attention on ImageNet~\cite{deng2009imagenet} for vision models and a subset of OpenWebText~\cite{Gokaslan2019OpenWeb} for language models. For evaluation on visual classification, we follow the protocols in ~\cite{jia2022visual} and evaluate on FGVC and VTAB-1k~\cite{zhai2019large}. FGVC contains 5 datasets of fine-grained natural image classification, each with around 10k training images. VTAB has 18 classification tasks that span natural image classification, specialized image classification (satellite, medical, \etc), and image structure understanding (\eg, object counting, depth estimation), with each task containing 1k training images. For evaluation on language generation, we compare to Stanford Alpaca~\cite{alpaca} by training on the same Alpaca dataset which contains 52k instruction-following data, and compare to Vicuna by training on an open-source dataset collected from ShareGPT conversations\footnote{\url{https://huggingface.co/datasets/anon8231489123/ShareGPT_Vicuna_unfiltered}}.

\minisection{Experimental setup}. We compare with several baselines for transfer learning: (i) \textbf{Linear} freezes the pre-trained backbone and only tunes a linear head on top of it, (ii) \textbf{Fully fine-tuning} tunes the whole backbone, (iii) \textbf{VPT}~\cite{jia2022visual} adds additional prompt tokens into the input as well as each intermediate layer and only tunes the prompt tokens, (iv) \textbf{LoRA}~\cite{hu2021lora} adds low-rank matrices onto the linear transform weights in the network and only tunes the low-rank matrices. The pre-trained backbone for visual classification is by default ViT-B~\cite{dosovitskiy2020image} pre-trained on ImageNet-1k. To align with the literature, we also test the performance on VTAB-1k using a ViT-B pre-trained on ImageNet-21k. For language generation, we use LLaMA-7B and LLaMA-13B~\cite{touvron2023llama} as the pre-trained backbones.

\subsection{Understanding Attention Refocusing in \method}
\label{sec:exp_understand}

\begin{figure}[t]
  \centering
  \includegraphics[width=1\linewidth]{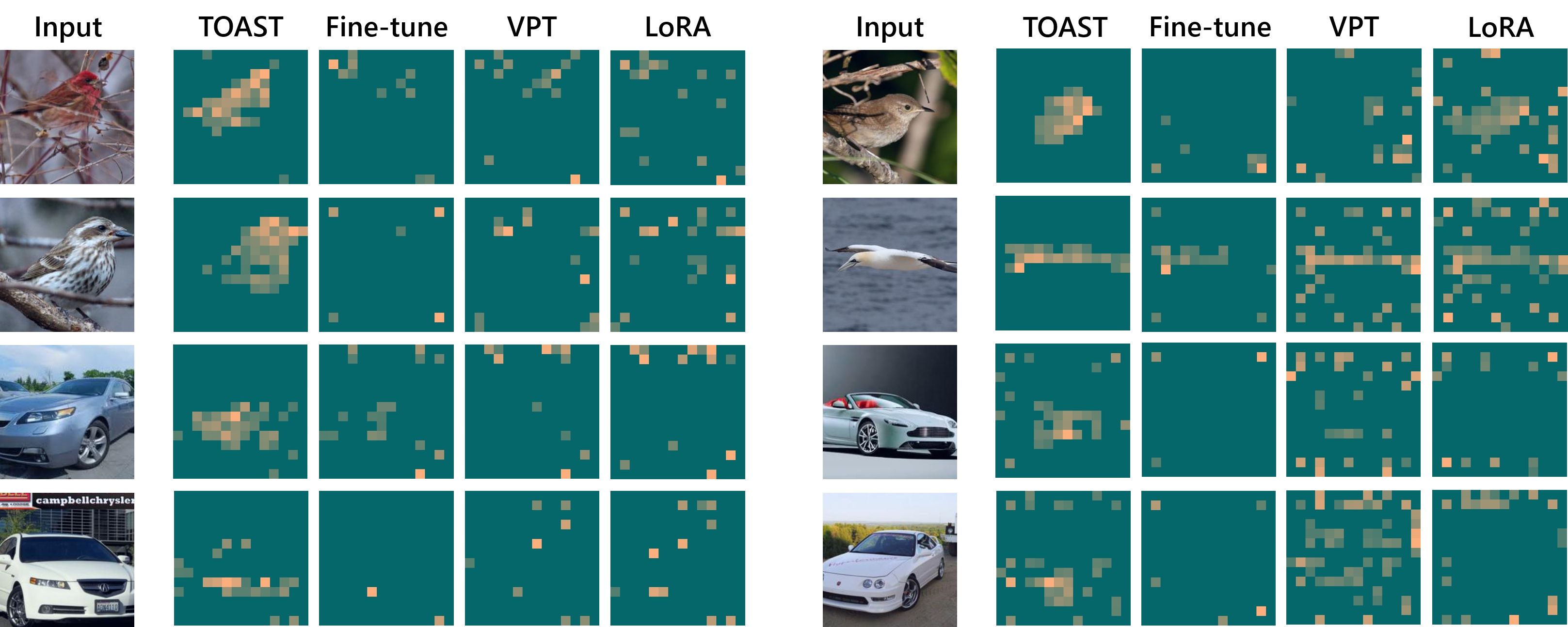}
  \caption{Comparison between the attention map of different models. The first two rows are evaluated on bird classification, and the last two on car classification. The attention of fine-tuning, LoRA, and VPT is noisy, while \method has cleaner attention that is focused on the task-relevant signals such as the foreground birds or the headlights and the badge of the cars.}
  \label{fig:att_compare}
  \vspace{-1em}
\end{figure}

\begin{wrapfigure}{r}{0.45\textwidth}
  \vspace{-2em}
  \begin{center}
      \includegraphics[width=0.9\linewidth]{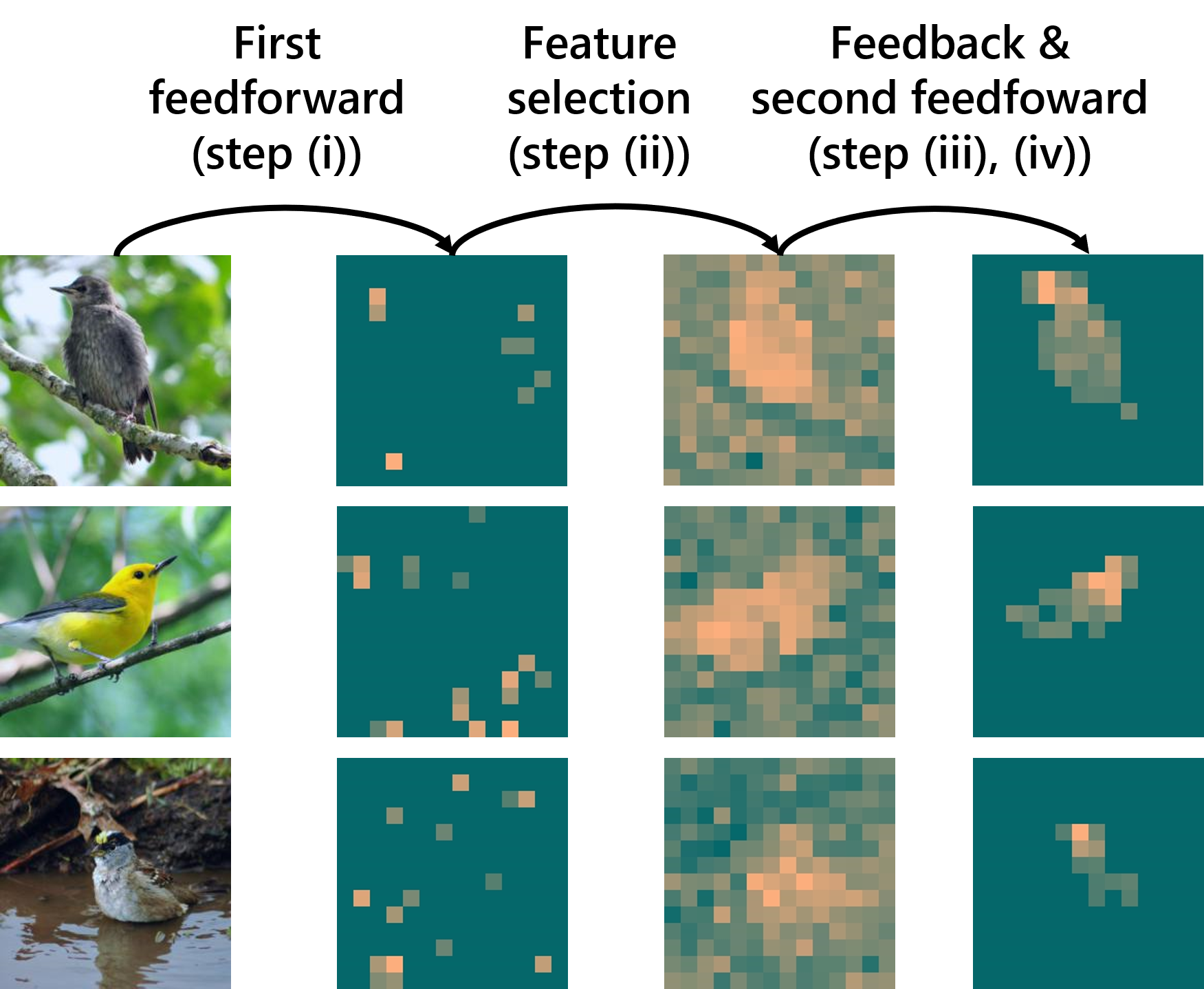}
  \end{center}
  \caption{Visualization of the attention maps during each step of model inference. The attention is extremely noisy in the first feedforward. The feature selection step coarsely selects the task-relevant features, and in the second feedforward, the attention is refined and refocused on the task-relevant objects.}
  \label{fig:att_refocus}
  \vspace{-2em}
\end{wrapfigure}

To understand how \method adapts to downstream tasks by refocusing its attention, we visualize how the attention changes during the inference of the top-down attention model (see Section~\ref{sec:method_arch}). In Figure~\ref{fig:att_refocus}, we show the attention map in the first feedforward pass, the similarity map in the feature selection step, as well as the attention in the second feedforward pass. We take FGVC bird classification as our example. One can see that in the first feedforward the pre-trained model fails to concentrate on the task-relevant objects. This explains to some extent why simply training a linear probing layer on top of the pre-trained backbone gives poor performance (see Section~\ref{sec:exp_vision}). \method addresses this problem with two stages. First, it selects the task-relevant features with the feature selection module. We can observe from the cosine similarity map that it coarsely selects the task-relevant objects. Then the reweighted tokens are sent back to the network to enhance the task-relevant features in the second feedforward run. We can see the attention is refined and refocused on the birds, providing better representations for the downstream task.

From a similar perspective, we explain why \method gives superior performance than fine-tuning as well as other baselines by comparing their attention maps. We use examples from FGVC bird and car classification. As shown in Figure~\ref{fig:att_compare}, for bird classification, \method clearly focuses on the foreground birds while other methods either have noisy attention or completely ignore the foreground object. On car classification, we observe that \method tends to concentrate on the headlights and the badge of the car which helps discriminate different brands of cars, while other methods have less explainable and noisier attention.

\subsection{Evaluation on Visual Classification}
\label{sec:exp_vision}

\begin{table}[t]
  \caption{Results on FGVC fine-grained classification. \method is able to outperform previous baselines by a large margin on different tasks and achieves state-of-the-art average performance.}
  \label{tab:fgvc}
  \vspace{1em}
  \centering
  \begin{small}
  \begin{tabular}{lcccccc}
    \toprule
     & CUB & Bird & Flower & Dog & Car & \textit{Avg} \\
    \midrule
    
    Linear & 76.8 & 47.3 & 81.7 & \textbf{97.7} & 60.3 & \textit{72.8} \\
    Fine-tune & 80.5 & 60.2 & 86.9 & 94.7 & 83.2 & \textit{81.1} \\
    VPT & 76.9 & 72.2 & 80.6 & 97.3 & 62.8 & \textit{78.0} \\
    LoRA & 82.5 & 71.2 & 81.2 & 97.5 & 76.6 & \textit{79.8} \\
    \rowcolor{lightgray!50}\method & \textbf{85.0} & \textbf{75.2} & \textbf{88.7} & 97.4 & \textbf{84.9} & \textbf{\textit{86.2}} \\

    \bottomrule
  \end{tabular}
  \end{small}
  \vspace{1em}
\end{table}

\begin{table*}[t]

\centering
\caption{Results on VTAB-1K benchmark. \method outperforms previous baselines on 11 out of 18 tasks for ImageNet-1k pre-trained model and 10 out of 18 tasks for ImageNet-21k pre-trained model. All methods are implemented in the same environment.}
\label{tab:vtab}
\vspace{0.5em}
\setlength{\tabcolsep}{0.3pt}
\resizebox{\textwidth}{!}{
\begin{tabular}{p{2.2cm}<{}|p{0.75cm}<{\centering}p{0.75cm}<{\centering}p{0.75cm}<{\centering}p{0.75cm}<{\centering}p{0.75cm}<{\centering}p{0.75cm}<{\centering}p{0.75cm}<{\centering}|p{0.75cm}<{\centering}p{0.75cm}<{\centering}p{0.75cm}<{\centering}p{0.75cm}<{\centering}|p{0.75cm}<{\centering}p{0.75cm}<{\centering}p{0.75cm}<{\centering}p{0.75cm}<{\centering}p{0.75cm}<{\centering}p{0.75cm}<{\centering}p{0.75cm}<{\centering}p{0.75cm}<{\centering}}
\toprule[1.5pt]
\multicolumn{1}{c|}{}&\multicolumn{7}{c|}{\textbf{Natural}}&\multicolumn{4}{c|}{\textbf{Specialized}}&\multicolumn{8}{c}{\textbf{Structured}}\\
&\multicolumn{1}{c}{{\rotatebox[origin=c]{90}{Cifar100}}}
&\multicolumn{1}{c}{{\rotatebox[origin=c]{90}{Caltech101}}}
&\multicolumn{1}{c}{{\rotatebox[origin=c]{90}{DTD}}}
&\multicolumn{1}{c}{{\rotatebox[origin=c]{90}{Flower102}}}
&\multicolumn{1}{c}{{\rotatebox[origin=c]{90}{Pets}}}
&\multicolumn{1}{c}{{\rotatebox[origin=c]{90}{SVHN}}}
&\multicolumn{1}{c|}{{\rotatebox[origin=c]{90}{Sun397}}}
&\multicolumn{1}{c}{{\rotatebox[origin=c]{90}{Camelyon}}}
&\multicolumn{1}{c}{{\rotatebox[origin=c]{90}{EuroSAT}}}
&\multicolumn{1}{c}{{\rotatebox[origin=c]{90}{Resisc45}}}
&\multicolumn{1}{c|}{{\rotatebox[origin=c]{90}{Retinopathy}}}
&\multicolumn{1}{c}{{\rotatebox[origin=c]{90}{Clevr-Count}}}
&\multicolumn{1}{c}{{\rotatebox[origin=c]{90}{Clevr-Dist}}}
&\multicolumn{1}{c}{{\rotatebox[origin=c]{90}{DMLab}}}
&\multicolumn{1}{c}{{\rotatebox[origin=c]{90}{KITTI-Dist}}}
&\multicolumn{1}{c}{{\rotatebox[origin=c]{90}{dSpr-Loc}}}
&\multicolumn{1}{c}{{\rotatebox[origin=c]{90}{dSpr-Ori}}}
&\multicolumn{1}{c}{{\rotatebox[origin=c]{90}{sNORB-Azim}}}
&\multicolumn{1}{c}{{\rotatebox[origin=c]{90}{sNORB-Ele}}}\\
\specialrule{0em}{1pt}{1pt}
\hline
\specialrule{0em}{1pt}{1pt}
\multicolumn{20}{l}{\emph{ImageNet-1k pre-trained}}\\
\hline
\specialrule{0em}{1pt}{1pt}
Fine-tune & 44.7 & 77.3 & 55.5 & 74.5 & 86.0 & \textbf{85.1} & 17.4 & \textbf{84.9} & 95.0 & 82.8 & 74.2 & 60.2 & 53.1 & 33.5 & 77.6 & 61.9 & 39.0 & 15.0 & 36.6 \\
VPT & 65.3 & 90.5 & 67.7 & 88.3 & 88.6 & 82.2 & 40.6 & 82.3 & 94.5 & 83.1 & 74.0 & 51.5 & 51.1 & 44.1 & 69.3 & 63.8 & 49.5 & 25.3 & 28.6 \\
LoRA & 69.3 & 88.8 & 66.6 & 90.3 & \textbf{90.3} & 81.9 & 41.5 & 83.4 & 94.8 & 83.5 & \textbf{75.0} & \textbf{66.8} & 56.9 & \textbf{48.9} & 77.6 & 76.2 & \textbf{53.5} & 26.6 & 37.1 \\
\rowcolor{lightgray!50}\method & \textbf{73.8} & \textbf{92.1} & \textbf{68.7} & \textbf{93.0} & 89.0 & 76.3 & \textbf{41.9} & 82.8 & \textbf{95.3} & \textbf{85.7} & 74.6 & 61.2 & \textbf{58.7} & 43.5 & \textbf{78.8} & \textbf{86.1} & 51.2 & \textbf{27.0} & \textbf{43.4} \\
\hline
\specialrule{0em}{1pt}{1pt}
\multicolumn{20}{l}{\emph{ImageNet-21k pre-trained}}\\
\hline
\specialrule{0em}{1pt}{1pt}
Fine-tune & 70.2 & 85.8 & 64.3 & 97.5 & 85.8 & \textbf{85.9} & 40.0 & 78.2 & 95.7 & 83.8 & 73.9 & 53.1 & 57.3 & 37.5 & 68.2 & 60.5 & 35.2 & 18.8 & 28.0 \\
VPT & 75.4 & 88.7 & 66.3 & 98.1 & 87.3 & 73.7 & 52.3 & 80.3 & 93.5 & 83.4 & 74.1 & 49.6 & 58.1 & 41.9 & 62.7 & 65.1 & 42.9 & 24.0 & 24.2 \\
LoRA & \textbf{83.6} & 89.4 & 66.2 & 98.6 & 89.4 & 83.8 & 52.6 & 81.1 & \textbf{95.8} & 84.6 & \textbf{74.7} & \textbf{77.6} & 59.5 & \textbf{46.8} & 74.1 & 73.0 & \textbf{48.6} & \textbf{25.6} & 32.2 \\
\rowcolor{lightgray!50}\method & 82.1 & \textbf{90.5} & \textbf{70.5} & \textbf{98.7} & \textbf{89.7} & 71.9 & \textbf{53.3} & \textbf{84.3} & 95.5 & \textbf{85.5} & 74.2 & 75.4 & \textbf{60.8} & 44.7 & \textbf{77.5} & \textbf{73.9} & 47.5 & 24.5 & \textbf{33.7} \\
\bottomrule[1.5pt]
\end{tabular}
}
\end{table*}

We evaluate \method on FGVC which contains 5 datasets of fine-grained natural image classification (Table~\ref{tab:fgvc}). We can see that \method outperforms fully fine-tuning as well as other baselines by a large margin. Especially, \method improves the average accuracy by 5\% over fine-tuning while training less parameters. This indicates an over-fitting issue in fine-tuning, which is also observed in VTAB-1k experiments. %
On the other hand, LoRA and VPT fail to improve over fine-tuning, posing a trade-off between parameter efficiency and downstream performance. Overall, \method is the only method that improves both downstream performance and parameter efficiency over fine-tuning.

We also test on VTAB, which contains 18 datasets of image classification or image structure understanding. Since each dataset only has 1k training images, VTAB simulates a setting with higher data scarcity. Results are shown in Table~\ref{tab:vtab}. Overall, \method reaches competitive performances, outperforming other baselines in 11 of 18 datasets for ImageNet-1k pre-trained model and 10 of 18 datasets for ImageNet-21k pre-trained model. We notice that \method shows more advantages on natural image classification than on specialized image classification or image structure understanding. This is possibly because the last two task categories have too large gaps from the pre-training task of natural image classification, and the features relevant to downstream tasks are absent in the pre-trained backbone. In this case, only refocusing the attention is not enough and tuning the feedforward backbone is essential for learning the features. %

\subsection{Evaluation on Language Generation}
\label{sec:exp_language}

\begin{figure}[t]
  \centering
  \includegraphics[width=0.95\linewidth]{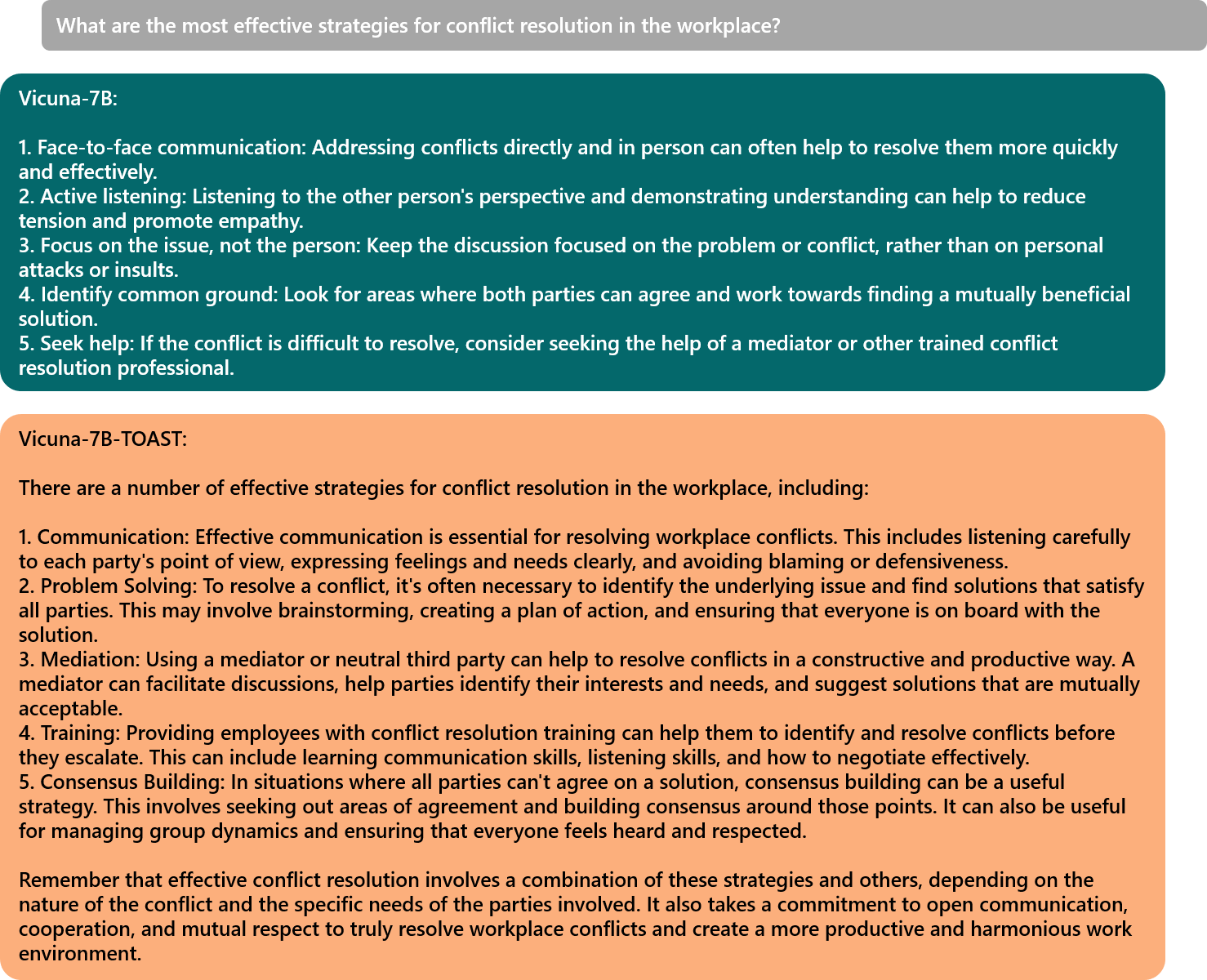}
  \caption{The responses given the question \emph{What are the most effective strategies for conflict resolution in the workplace.} Both the fine-tuned baseline and \method list a few suggestions for the question, while \method provides more detailed explanations for each suggestion. 
  }
  \label{fig:llama1}
\end{figure}

\begin{figure}[t]
  \centering
  \includegraphics[width=0.95\linewidth]{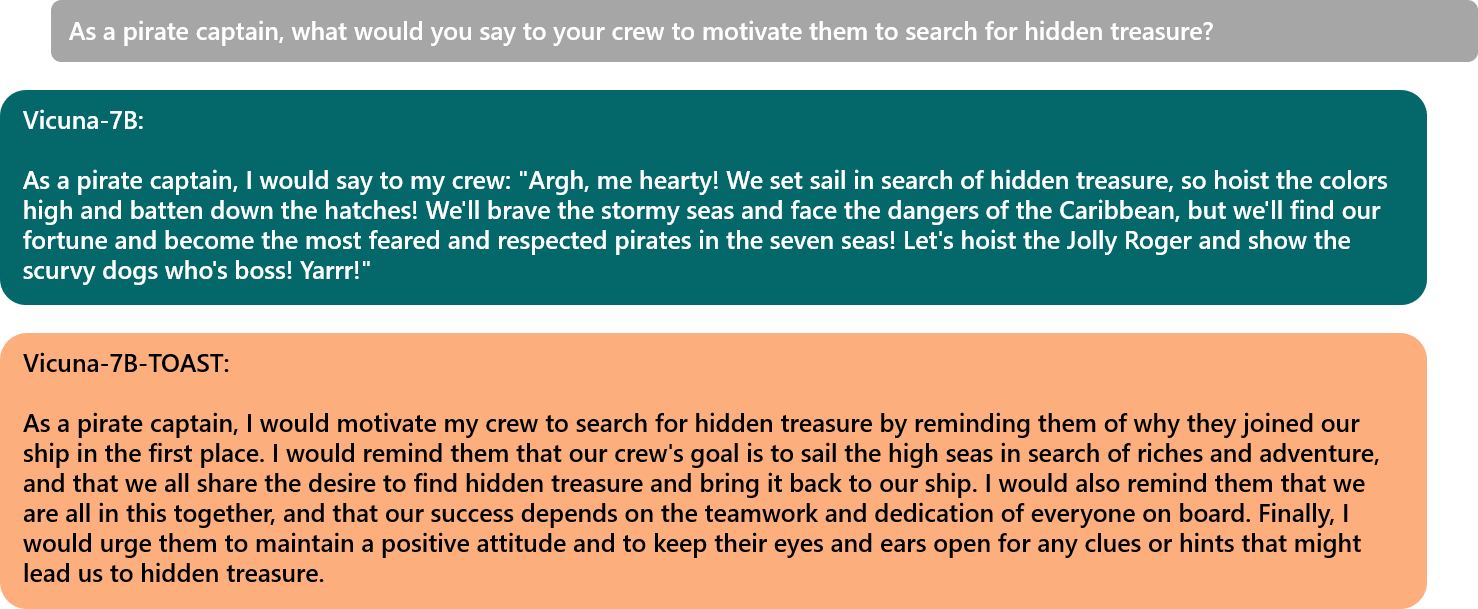}
  \caption{The responses given the question \emph{As a pirate captain, what would you say to your crew to motivate them to search for hidden treasure}. \method takes a motivational approach and provides a detailed description of the prepared speech. On the other hand, the fine-tuned baseline directly embraces the private persona and gives a vivid and engaging answer.
  }
  \label{fig:llama2}
\end{figure}

We apply \method to LLaMA-7B and LLaMA-13B~\cite{touvron2023llama} and tune it on the Alpaca dataset~\cite{alpaca} to compare to Stanford Alpaca, and on the open-source ShareGPT dataset to compare to Vicuna~\cite{vicuna2023}. For evaluation, we follow the pipeline in Vicuna~\cite{vicuna2023}, \ie, we ask questions that span different categories and use GPT-4~\cite{openai2023gpt4} to score the answers provided by the model. Scores are based on the helpfulness, relevance, accuracy, and level of detail of the answers. In Table~\ref{tab:llama}, we compare the scores of different models on each category of questions. All the scores are on a scale of 1-10. Overall, \method has slightly better performance than fine-tuning for each baseline model, while LoRA is less competitive. We observe that \method is better at providing detailed and informed answers, which makes it a better assistant for questions about general knowledge (Generic, Knowledge, Common). Figure~\ref{fig:llama1} is an example. Given the question \textit{What are the most effective strategies for conflict resolution in the workplace}, while the fine-tuned baseline lists several suggestions, \method is able to provide more detailed explanations for each suggestion. On the other hand, fine-tuning is more helpful when following the instruction for creative questions such as Roleplay and Counterfact. Figure~\ref{fig:llama2} gives an example of a roleplay question. \method provides a detailed explanation of what to say to the crews, while the fine-tuned baseline directly creates a short speech which more adheres to the instruction. These observations are consistent with the underlying mechanism of \method and fine-tuning, \ie, \method does not modify the pre-trained backbone and thus is able to ``remember'' all the knowledge learned during pre-training, meanwhile fine-tuning may forget what is learned during pre-training when modifying the weights but is able to better follow the instruction in this way. More examples are shown in Appendix.

\subsection{\method Is Adaptable to Different Model Architectures and Tasks}
\label{sec:exp_versatile}

\minisection{\method is adaptable to Convnets}. In previous experiments, we use Transformer as the backbone. We show that we can also use \method on convolutional networks (Convnets). First, we need to design a top-down attention module for Convnets: (i) we keep the design of the feature selection module, (ii) we change the linear layers in the feedback path into deconvolutional layers so that the bottom-up and top-down signals in each layer have the same shape, (iii) since there is no self-attention in Convnets, we directly add the top-down signal onto the bottom-up input of each convolutional layer. Then the pre-tuning and tuning stages are the same as the transformer setting. In our experiments, we choose ConvNeXt~\cite{liu2022convnet} as the backbone and test on FGVC (Table~\ref{tab:fgvc_convnext}). We observe similar results as in Transformer that \method has superior performance than fine-tune and LoRA. This implies attention refocusing is also important for Convnets.

\minisection{\method is adaptable to larger models.} To see if \method can scale to larger models, we test ViT-L pre-trained on ImageNet-21k. As shown in Table~\ref{tab:fgvc_in21k}, on the larger model, \method still delivers the best performance. An interesting observation is that LoRA is able to outperform fine-tune in this setting, possibly because the pre-trained representation is strong and general enough and largely modifying the backbone to learn new features is not necessary. 

\begin{table}[t]
  \caption{Results of Instruction Tuning. We use GPT-4 to evaluate the performance. LoRA has inferior performance while \method is able to outperform fully fine-tuned Alpaca and Vicuna. $^\dagger$We train the Alpaca and Vicuna baselines using the same open-source data as LoRA and \method and report the performance thereof.}
  \label{tab:llama}
  \vskip 0.1in
  \centering
  \begin{small}
  \begin{tabular}{lcccccccc}
    \toprule
      & Generic & Knowledge & Roleplay & Common & Fermi & Counterfact & Writing & \textit{Avg} \\
    \midrule
     Alpaca-7B$^\dagger$ & 7.8 & 8.5 & \textbf{8.0} & \textbf{8.0} & 4.3 & \textbf{8.7} & \textbf{9.7} & \textit{7.9}\\
     \ - \ LoRA & 6.7 & 7.3 & 6.7 & 7.0 & 5.0 & 7.3 & 8.0 & \textit{6.9}\\
     \rowcolor{lightgray!50} \ - \ \method & \textbf{8.0} & \textbf{9.0} & 7.7 & \textbf{8.0} & \textbf{7.0} & 8.0 & 8.7 & \textbf{\textit{8.1}} \\
     \midrule
     Vicuna-7B$^\dagger$ & 8.3 & 8.8 & 8.2 & 8.0 & \textbf{6.7} & \textbf{7.7} & 8.8 & \textit{8.1}\\
     \rowcolor{lightgray!50} \ - \ \method & \textbf{8.7} & \textbf{9.0} & \textbf{8.7} & \textbf{9.0} & 6.5 & 7.0 & \textbf{9.0} & \textbf{\textit{8.3}} \\
     \midrule
     Vicuna-13B$^\dagger$ & 7.6 & 8.5 & \textbf{9.3} & 8.2 & \textbf{7.0} & \textbf{8.0} & 8.7 & \textit{8.2}\\
     \rowcolor{lightgray!50} \ - \ \method & \textbf{8.9} & \textbf{9.0} & 8.0 & \textbf{9.0} & 6.7 & \textbf{8.0} & \textbf{9.0} & \textbf{\textit{8.4}} \\

    \bottomrule
  \end{tabular}
  \end{small}
\end{table}

\setlength{\tabcolsep}{5pt}
\begin{table}[t]
\parbox{.45\linewidth}{
  \caption{Results on FGVC with ConvNeXt-B backbone.}
  \label{tab:fgvc_convnext}
  \vskip 0.1in
  \centering
  \begin{scriptsize}
  \begin{tabular}{lcccccc}
    \toprule
     & CUB & Birds & Flower & Dogs & Cars & \textit{Avg} \\
    \midrule
    
    Fine-tune &  87.5 & 72.3 & 97.1 & 86.3 & \textbf{87.7} & \textit{86.2} \\
    LoRA & 89.6 & 75.8 & \textbf{99.3} & \textbf{88.5} & 67.7 & \textit{84.2} \\
    \rowcolor{lightgray!50}\method & \textbf{90.2} & \textbf{85.6} & \textbf{99.2} & \textbf{88.4} & 85.8 & \textbf{\textit{89.8}} \\

    \bottomrule
  \end{tabular}
  \end{scriptsize}
}
\hfill
\parbox{.45\linewidth}{
  \caption{Results on FGVC with ViT-L backbone pre-trained on ImageNet-21k.}
  \label{tab:fgvc_in21k}
  \vskip 0.1in
  \centering
  \begin{scriptsize}
  \begin{tabular}{lcccccc}
    \toprule
     & CUB & Bird & Flower & Dog & Car & \textit{Avg} \\
    \midrule
    
    Fine-tune & 88.3 & 69.4 & 98.1 & 89.8 & 84.3 & 86.0 \\
    LoRA & 89.1 & 73.9 & 98.2 & \textbf{94.3} & 78.1 & 86.7 \\
    \rowcolor{lightgray!50}\method & \textbf{89.5} & \textbf{75.4} & \textbf{98.5} & 93.4 & \textbf{85.3} & \textbf{\textit{88.4}} \\

    \bottomrule
  \end{tabular}
  \end{scriptsize}
}
\end{table}
\setlength{\tabcolsep}{6pt}

\begin{table}[t]
\parbox{.47\linewidth}{
  \caption{Results on Semantic Segmentation. \method consistently outperforms LoRA and VPT but still lags behind fully fine-tuning.}
  \label{tab:seg}
  \vspace{1em}
  \centering
  \begin{small}
  \begin{tabular}{lcc}
    \toprule
     & PASCAL VOC & ADE20K \\
    \midrule
    
    Fine-tune & \textbf{82.05} & \textbf{47.89}  \\
    VPT & 76.80 & 41.42 \\
    LoRA & 78.43 & 42.94 \\
    \rowcolor{lightgray!50}\method & 80.44 & 45.11 \\

    \bottomrule
  \end{tabular}
  \end{small}
}
\hfill
\parbox{.44\linewidth}{
  \caption{Ablation studies on the pre-tuning stage, the token-wise and channel-wise attention in \method.}
  \label{tab:ablation}
  \vspace{1em}
  \centering
  \begin{small}
  \begin{tabular}{lc}
    \toprule
    Model & FGVC Avg Acc \\
    \midrule
    \method & \textbf{86.2} \\
    w/o pre-tuning & 81.9 \\
    w/o token att & 82.8 \\
    w/o channel att & 74.7 \\

    \bottomrule
  \end{tabular}
  \end{small}
}
\vspace{-1em}
\end{table}

\minisection{\method is adaptable to semantic segmentation}. Previous work~\cite{jia2022visual} shows that PEFT methods are not comparable to fine-tuning on dense prediction tasks such as semantic segmentation. Here we test \method on semantic segmentation on two datasets, PASCAL VOC~\cite{pascal-voc-2012} and ADE20K~\cite{zhou2017scene}. We use ImageNet-21k pre-trained ViT-B as the backbone and UperNet~\cite{xiao2018unified} as the segmentation head. Since segmentation requires the model to encode low-level visual information, we find that feedback from the middle layer instead of the last layer gives better performance. From Table~\ref{tab:seg}, we observe that \method has better performance than VPT and LoRA, although still underperforms fine-tuning. One possible reason is that the backbone is pre-trained on image classification which has too large a discrepancy with segmentation tasks in terms of the hierarchy and semantics of the required visual representations.%

\subsection{Exploring Parameter-Efficient \method}
\label{sec:exp_pe}

\begin{wraptable}{r}{0.5\textwidth}
    \vspace{-1em}
  \caption{Evaluation of \methodlite on FGVC visual classification and Alpaca language generation. \methodlite outperforms LoRA and VPT with a similar number of tunable parameters.}
  \label{tab:pe}
  \centering
  \begin{small}
  \begin{tabular}{lcccc}
    \toprule
     & \multicolumn{2}{c}{FGVC} & \multicolumn{2}{c}{Alpaca} \\
     & \#Param & Acc & \#Param & Score \\
    \midrule
    Fine-tune & 87M & 81.1 & 7B & 7.9\\
    \midrule
    LoRA & 0.3M & 79.8 & 4.2M & 6.9\\
    VPT & 0.9M & 78.0 & - & - \\
    \midrule
    \method & 14M & \textbf{86.2} & 537M & \textbf{8.1} \\
    \methodlite & 0.9M & 86.0 & 19M & 7.4 \\

    \bottomrule
  \end{tabular}
  \end{small}
\end{wraptable}

In Section~\ref{sec:method_toast} we mention \method is tuning around 15\% of the parameters and most of the tunable parameters are from the feedback path. To further reduce the number of tunable parameters and match the parameter efficiency of methods such as LoRA and VPT, we propose \methodlite which applies LoRA on the feedback path. In this way, only less than 1\% of the parameters are tuned. Here we evaluate the performance of \methodlite on FGVC and Alpaca. Results are shown in Table~\ref{tab:pe}. We can see that although \methodlite tunes much fewer parameters than \method, it performs on par with \method on FGVC. It also largely outperforms LoRA and VPT while having a similar level of parameter efficiency. For Alpaca, \methodlite has a degraded performance compared to \method but still outperforms LoRA, making it a strong baseline for Parameter-Efficient Fine-Tuning. See Appendix for more results of \methodlite.

\subsection{Ablation Studies}
\label{sec:exp_ablation}

We conduct ablation studies to show the importance of several designs of \method: (i) the pre-tuning stage which provides a better initialization of the top-down attention module, and (ii) the token-wise and channel-wise attention in the top-down attention module. For each ablation, we remove the pre-tuning stage, remove the token selection in the feature selection module, and freeze the channel selection as well as the feedback path, respectively. Note that we freeze the feedback path because it contains linear transforms on the channel dimension and thus also plays a role in channel selection. Results are shown in Table~\ref{tab:ablation}. First, we observe that \method without pre-tuning has a considerable performance drop from 86.2\% to 81.9\%. This indicates a proper initialization of the top-down attention module is crucial. Notably, \method without pre-tuning still outperforms fine-tuning, proving the effectiveness of attention refocusing. Second, we can see that removing the token-wise attention or channel-wise attention will both harm the performance. Specifically, removing channel-wise attention has a larger impact, indicating that at the same position in an image, the pre-trained model is usually not focusing on the features concerned by downstream tasks.
\section{Current Limitations of \method}

Despite the promising performances, the major drawback of \method and \methodlite is the computation overhead since the feedforward path is run twice, which approximately doubles the FLOPS of the model. Although it is worth noting that even though doubling the FLOPS, \method is still able to outperform fine-tuned ViT-L which has twice as many FLOPS as \method (Table~\ref{tab:appendix_fgvc_early_late_feedback}). To further improve computational efficiency, we seek ways to avoid running the feedforward path twice. Specifically, we try out two different design choices, early feedback, and late feedback. Early feedback means feedback from the middle layer instead of the last layer. In this way, we can only run the layers before the middle layer in the first feedforward. Late feedback means feedback from the last layer to the middle layer instead of the first layer. In this way, we can share the outputs before the middle layer in two feedforward runs since they do not receive any feedback and thus have the same outputs in both feedforward runs. We test these two designs on FGVC (Table~\ref{tab:appendix_fgvc_early_late_feedback}). We observe that early and late feedback reduces the FLOPS at the cost of slightly degrading the performance. However, we also find this is not always the case. For example, for semantic segmentation, we find that early feedback is actually better than regular feedback (45.11\ \emph{vs.}\ 43.59 mIoU on ADE20K), probably because segmentation requires more fine-grained and low-level features and the middle layer contains more low-level information than the last layer.

\begin{table}[t]
\vspace{1em}
  \caption{Results on FGVC with early or late feedback. All models are ViT-B unless noted otherwise. Vanilla \method doubles the FLOPS over fine-tuned ViT-B although it still outperforms fine-tuned ViT-L which has twice as many FLOPS as \method. The early and late feedback models further reduce the FLOPS but with a cost of degrading performances compared to \method. }
  \label{tab:appendix_fgvc_early_late_feedback}
  \vskip 0.1in
  \centering
  \begin{small}
  \begin{tabular}{lccccccc}
    \toprule
     & FLOPS & CUB & Birds & Flower & Dogs & Cars & \textit{Avg} \\
    \midrule
    Fine-tune & 1x & 80.5 & 60.2 & 86.9 & 94.7 & 83.2 & \textit{81.1} \\
    \textcolor{gray}{Fine-tune (ViT-L)} & \textcolor{gray}{4x} & \textcolor{gray}{88.3} & \textcolor{gray}{69.4} & \textcolor{gray}{98.1} & \textcolor{gray}{89.8} & \textcolor{gray}{84.3} & \textcolor{gray}{\textit{86.0}} \\
    \midrule
    \method & 2x & \textbf{85.0} & 75.2 & \textbf{88.7} & \textbf{97.4} & \textbf{84.5} & \textbf{\textit{86.2}} \\
    \method-Early & 1.5x & 84.2 & 74.0 & 85.2 & 97.2 & 81.7 & 84.5 \\
    \method-Late & 1.5x & 83.7 & \textbf{75.9} & 86.1 & 97.3 & 77.5 & 84.1\\
    
    \bottomrule
  \end{tabular}
  \end{small}
\end{table}
\section{Conclusion}
This work is motivated by the empirical observation that previous transfer learning methods often fail to focus the model's attention on task-relevant signals, which possibly leads to suboptimal performance on downstream tasks. We show that refocusing attention is the key to better transfer learning performance. We propose Top-Down Attention Steering (\method) which transfers to a new task by steering the attention onto the task-specific features. Specifically, \method freezes the pre-trained backbone and tunes an additional top-down attention module on the downstream task to steer the attention. Compared to previous baselines, \method is able to achieve state-of-the-art results on fine-grained visual classification as well as instruction-following language generation while only tuning a small portion of the parameters. 

{\small
\bibliographystyle{plain}
\bibliography{egbib}
}

\newpage
\appendix
\section{Additional Implementation Details}

\subsection{Pre-Trained Backbone}

For the feedforward ViT-B backbone with ImageNet-1k pre-training, we use the implementation from DeiT~\cite{pmlr-v139-touvron21a} and pre-train on ImageNet-1k with the same recipe, \ie, using AdamW optimizer to pre-train for 300 epochs,
with a batch size of 512, a base learning rate of 5e-4, and 5 warm-up epochs. For the ViT-B and ViT-L backbone with ImageNet-21k pre-training, we take the checkpoints from HuggingFace\footnote{https://huggingface.co/google/vit-base-patch16-224-in21k} \footnote{https://huggingface.co/google/vit-large-patch16-224-in21k} and convert them into DeiT style. For ConvNeXt we directly borrow the implementation and checkpoints from the original GitHub repository\footnote{https://github.com/facebookresearch/ConvNeXt}. For LLaMA-7B and LLaMA-13B, we take the checkpoints provided by the community\footnote{https://huggingface.co/decapoda-research/llama-7b-hf} \footnote{https://huggingface.co/decapoda-research/llama-13b-hf}. 

\subsection{Pre-Tuning Stage}

For vision models such as ViT and ConvNeXt, we first add a randomly initialized top-down attention module onto the pre-trained backbone and then pre-tune the top-down attention module on ImageNet-1k classification. In this process, the feedforward backbone is frozen. We pre-tune the model for 30 epochs using the AdamW optimizer, with 3 warm-up epochs, and 3 cool-down epochs, a learning rate of 0.0005. We also disable the cutmix and mixup. Except for the supervised loss, we also add the variational loss~\cite{shi2023top} which encourages the feedback layer in $\ell$-th layer to reconstruct the input feature to $\ell$-th layer from its output. We set the weight of variational loss as 0.03.

For language models, we pre-tune on a subset of OpenWebText~\cite{Gokaslan2019OpenWeb}. The subset contains 200k lines sampled from the original dataset. We train for 1 epoch with a batch size of 32 and 4 gradient accumulation steps. We use a learning rate of 3e-5. We use DeepSpeed\footnote{https://github.com/microsoft/DeepSpeed} parameter offloading to avoid OOM errors. 

\subsection{Tuning Stage}

For FGVC experiments, we use the training recipe in \cite{jia2022visual}. Specifically, on each dataset, we use a learning rate of 0.01 for \method, LoRA, and VPT, and use 0.003 for fine-tuning. We use a batch size of 32. 

For VTAB experiments, we follow~\cite{jia2022visual} to do a grid search on the best learning rate and weight decay for each model and each dataset. Specifically, we take 800 images from the training set to train the model and use the rest 200 images for validation. We pick the set of hyperparameters that has the highest validation performance. Then we use the same hyperparameters to train the model on all 1000 images and test on the testing set. For each dataset, we run it five times with random seeds and report the average results. 

For experiments on Alpaca and Vicuna, we use the same training recipe as the Stanford Alpaca repository\footnote{https://github.com/tatsu-lab/stanford\_alpaca}. During the evaluation, we use a temperature of 0.7. The evaluation protocol follows the one in Vicuna~\cite{vicuna2023} except we sample 30 questions from the original list of 80 questions.

\section{Additional Results of \methodlite}

In this section, we provide more results on \methodlite and compare it to other PEFT algorithms such as LoRA and VPT. For visual classification, we show the results on FGVC in Table~\ref{tab:appendix_fgvc}. We can see the \methodlite has a similar number of tunable parameters as LoRA and VPT while obtaining better or comparable performances on all five datasets. It also outperforms fine-tuning while other PEFT methods fail to. We also show the results on VTAB-1k (Table~\ref{tab:appendix_vtab}). We observe that \methodlite normally is not able to match the performance of \method. \methodlite performs on par with LoRA on classification while has a worse performance on structure understanding.%

We also provide more results of \methodlite on language generation task. As shown in Table~\ref{tab:appendix_llama}, \methodlite has a downgraded performance compared to \method, but still outperforms LoRA. %

\begin{table}[t]
  \caption{Results on FGVC with \method and \methodlite. \methodlite is able to improve the performance by a large margin over LoRA and VPT while tuning a similar number of parameters.}
  \label{tab:appendix_fgvc}
  \vskip 0.1in
  \centering
  \begin{small}
  \begin{tabular}{lccccccc}
    \toprule
     & \# Params & CUB & Birds & Flower & Dogs & Cars & \textit{Avg} \\
    \midrule
    Linear & 0.2M & 76.8 & 47.3 & 81.7 & \textbf{97.7} & 60.3 & \textit{72.8} \\
    Fine-tune & 87M & 80.5 & 60.2 & 86.9 & 94.7 & 83.2 & \textit{81.1} \\
    VPT & 0.9M & 76.9 & 72.2 & 80.6 & 97.3 & 62.8 & \textit{78.0} \\
    LoRA & 0.3M & 82.5 & 71.2 & 81.2 & 97.5 & 76.6 & \textit{79.8} \\
    \rowcolor{lightgray!50}\method & 14M & \textbf{85.0} & 75.2 & 88.7 & \textbf{97.4} & \textbf{84.5} & \textbf{\textit{86.2}} \\
    \rowcolor{lightgray!50}\methodlite & 0.9M & 84.5 & \textbf{76.9} & \textbf{89.4} & \textbf{97.4} & 82.0 & \textit{86.0} \\
    
    \bottomrule
  \end{tabular}
  \end{small}
  \vskip 0.1in
\end{table}

\begin{table*}[t]

\centering
\caption{Results on VTAB-1K benchmark with \method and \methodlite.}
\label{tab:appendix_vtab}
\vspace{0.5em}
\setlength{\tabcolsep}{0.3pt}
\resizebox{\textwidth}{!}{
\begin{tabular}{p{2.0cm}<{}p{0.9cm}<{\centering}|p{0.75cm}<{\centering}p{0.75cm}<{\centering}p{0.75cm}<{\centering}p{0.75cm}<{\centering}p{0.75cm}<{\centering}p{0.75cm}<{\centering}p{0.75cm}<{\centering}|p{0.75cm}<{\centering}p{0.75cm}<{\centering}p{0.75cm}<{\centering}p{0.75cm}<{\centering}|p{0.75cm}<{\centering}p{0.75cm}<{\centering}p{0.75cm}<{\centering}p{0.75cm}<{\centering}p{0.75cm}<{\centering}p{0.75cm}<{\centering}p{0.75cm}<{\centering}p{0.75cm}<{\centering}}
\toprule[1.5pt]
\multicolumn{2}{c|}{}&\multicolumn{7}{c|}{\textbf{Natural}}&\multicolumn{4}{c|}{\textbf{Specialized}}&\multicolumn{8}{c}{\textbf{Structured}}\\
&\multicolumn{1}{c|}{{\rotatebox[origin=c]{90}{\# Params}}}
&\multicolumn{1}{c}{{\rotatebox[origin=c]{90}{Cifar100}}}
&\multicolumn{1}{c}{{\rotatebox[origin=c]{90}{Caltech101}}}
&\multicolumn{1}{c}{{\rotatebox[origin=c]{90}{DTD}}}
&\multicolumn{1}{c}{{\rotatebox[origin=c]{90}{Flower102}}}
&\multicolumn{1}{c}{{\rotatebox[origin=c]{90}{Pets}}}
&\multicolumn{1}{c}{{\rotatebox[origin=c]{90}{SVHN}}}
&\multicolumn{1}{c|}{{\rotatebox[origin=c]{90}{Sun397}}}
&\multicolumn{1}{c}{{\rotatebox[origin=c]{90}{Camelyon}}}
&\multicolumn{1}{c}{{\rotatebox[origin=c]{90}{EuroSAT}}}
&\multicolumn{1}{c}{{\rotatebox[origin=c]{90}{Resisc45}}}
&\multicolumn{1}{c|}{{\rotatebox[origin=c]{90}{Retinopathy}}}
&\multicolumn{1}{c}{{\rotatebox[origin=c]{90}{Clevr-Count}}}
&\multicolumn{1}{c}{{\rotatebox[origin=c]{90}{Clevr-Dist}}}
&\multicolumn{1}{c}{{\rotatebox[origin=c]{90}{DMLab}}}
&\multicolumn{1}{c}{{\rotatebox[origin=c]{90}{KITTI-Dist}}}
&\multicolumn{1}{c}{{\rotatebox[origin=c]{90}{dSpr-Loc}}}
&\multicolumn{1}{c}{{\rotatebox[origin=c]{90}{dSpr-Ori}}}
&\multicolumn{1}{c}{{\rotatebox[origin=c]{90}{sNORB-Azim}}}
&\multicolumn{1}{c}{{\rotatebox[origin=c]{90}{sNORB-Ele}}}\\
\specialrule{0em}{1pt}{1pt}
\hline
\specialrule{0em}{1pt}{1pt}
\multicolumn{21}{l}{\emph{ImageNet-1k pre-trained}}\\
\hline
\specialrule{0em}{1pt}{1pt}
Fine-tune & 87M & 44.7 & 77.3 & 55.5 & 74.5 & 86.0 & \textbf{85.1} & 17.4 & \textbf{84.9} & 95.0 & 82.8 & 74.2 & 60.2 & 53.1 & 33.5 & 77.6 & 61.9 & 39.0 & 15.0 & 36.6 \\
VPT & 0.9M & 65.3 & 90.5 & 67.7 & 88.3 & 88.6 & 82.2 & 40.6 & 82.3 & 94.5 & 83.1 & 74.0 & 51.5 & 51.1 & 44.1 & 69.3 & 63.8 & 49.5 & 25.3 & 28.6 \\
LoRA & 0.3M & 69.3 & 88.8 & 66.6 & 90.3 & \textbf{90.3} & 81.9 & 41.5 & 83.4 & 94.8 & 83.5 & \textbf{75.0} & \textbf{66.8} & 56.9 & \textbf{48.9} & 77.6 & 76.2 & \textbf{53.5} & 26.6 & 37.1 \\
\rowcolor{lightgray!50}\method & 14M & \textbf{73.8} & \textbf{92.1} & \textbf{68.7} & 93.0 & 89.0 & 76.3 & 41.9 & 82.8 & 95.3 & 85.7 & 74.6 & 61.2 & \textbf{58.7} & 43.5 & \textbf{78.8} & \textbf{86.1} & 51.2 & \textbf{27.0} & \textbf{43.4} \\
\rowcolor{lightgray!50}\methodlite & 0.9M & 69.0 & 91.0 & 65.9 & \textbf{93.3} & 88.0 & 78.9 & \textbf{42.9} & 81.6 & \textbf{96.1} & \textbf{86.0} & 74.3 & 62.8 & 53.3 & 44.8 & 73.7 & 66.0 & 41.2 & 26.5 & 39.0 \\
\bottomrule[1.5pt]
\end{tabular}
}
\end{table*}

\begin{table}[t]
  \caption{Results of Instruction Tuning on Alpaca. We use GPT-4 to evaluate the performance. LoRA has inferior performance while \method is able to outperform fully fine-tuned Alpaca.}
  \label{tab:appendix_llama}
  \vskip 0.1in
  \centering
  \begin{scriptsize}
      \begin{tabular}{lccccccccc}
    \toprule
      & \# Params & Generic & Knowledge & Roleplay & Common & Fermi & Counterfactual & Writing & \textit{Avg} \\
    \midrule
     Alpaca & 7B & 7.8 & 8.5 & \textbf{8.0} & \textbf{8.0} & 4.3 & \textbf{8.7} & \textbf{9.7} & \textit{7.9}\\
     LoRA & 4.2M & 6.7 & 7.3 & 6.7 & 7.0 & 5.0 & 7.3 & 8.0 & \textit{6.9}\\
     \rowcolor{lightgray!50}\method & 537M & \textbf{8.0} & 9.0 & 7.7 & \textbf{8.0} & \textbf{7.0} & 8.0 & 8.7 & \textbf{\textit{8.1}} \\
     \rowcolor{lightgray!50}\methodlite & 19M & 7.3 & \textbf{9.3} & 6.7 & 7.8 & 5.1 & 7.0 & 8.3 & 7.4 \\

    \bottomrule
  \end{tabular}
  \end{scriptsize}
  \vspace{-1em}
\end{table}

\section{Additional Results of Language Generation}

Figure 7-11 show additional results on language generation for Alpaca-7B, Vicuna-7B, and Vicuna-13B.

\begin{figure}[t]
  \centering
  \includegraphics[width=0.95\linewidth]{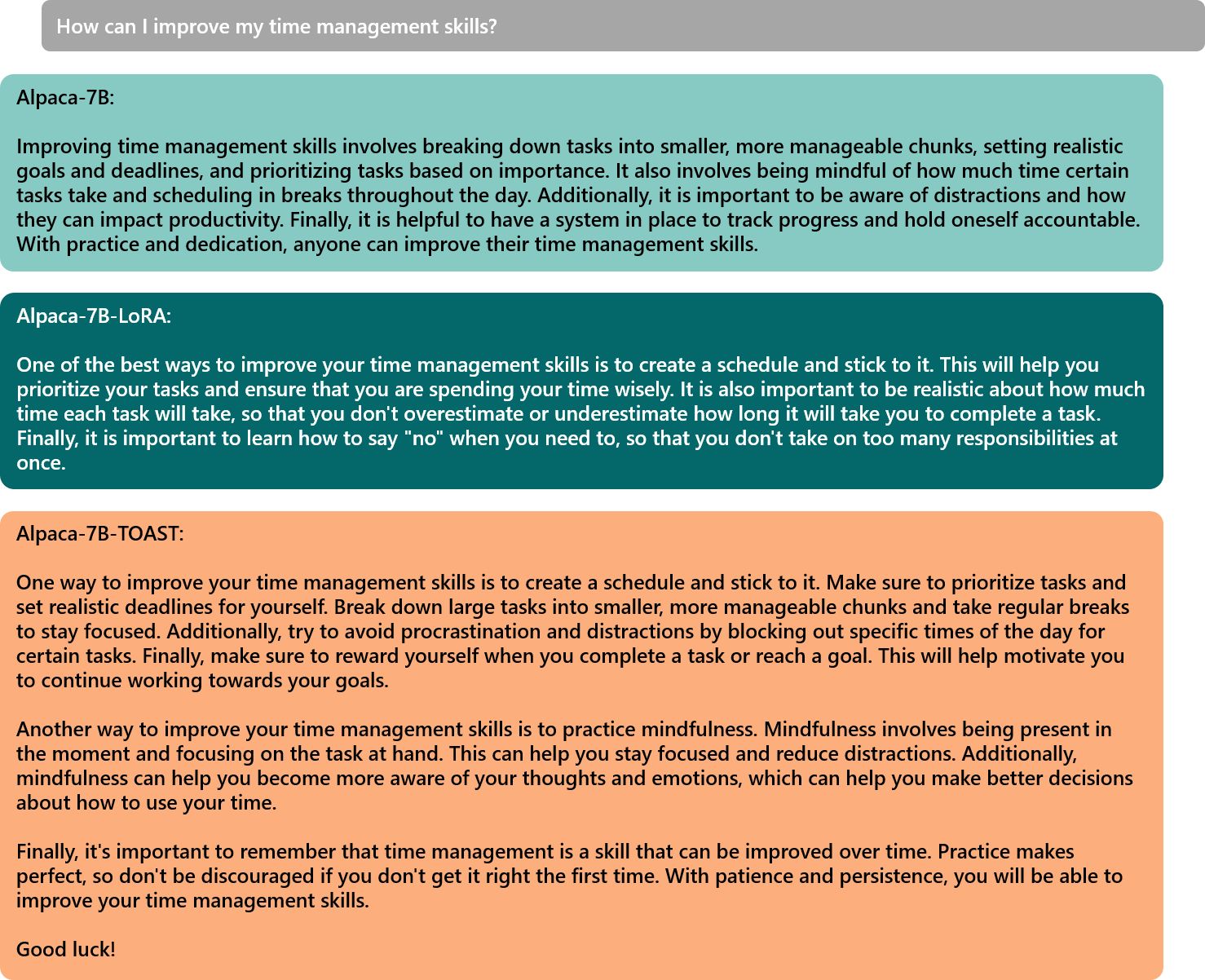}
  \caption{The responses given the question \emph{How can I improve my time management skills}. %
  }
  \label{fig:appendix_llama1}
  \vspace{-1em}
\end{figure}

\begin{figure}[t]
  \centering
  \includegraphics[width=0.95\linewidth]{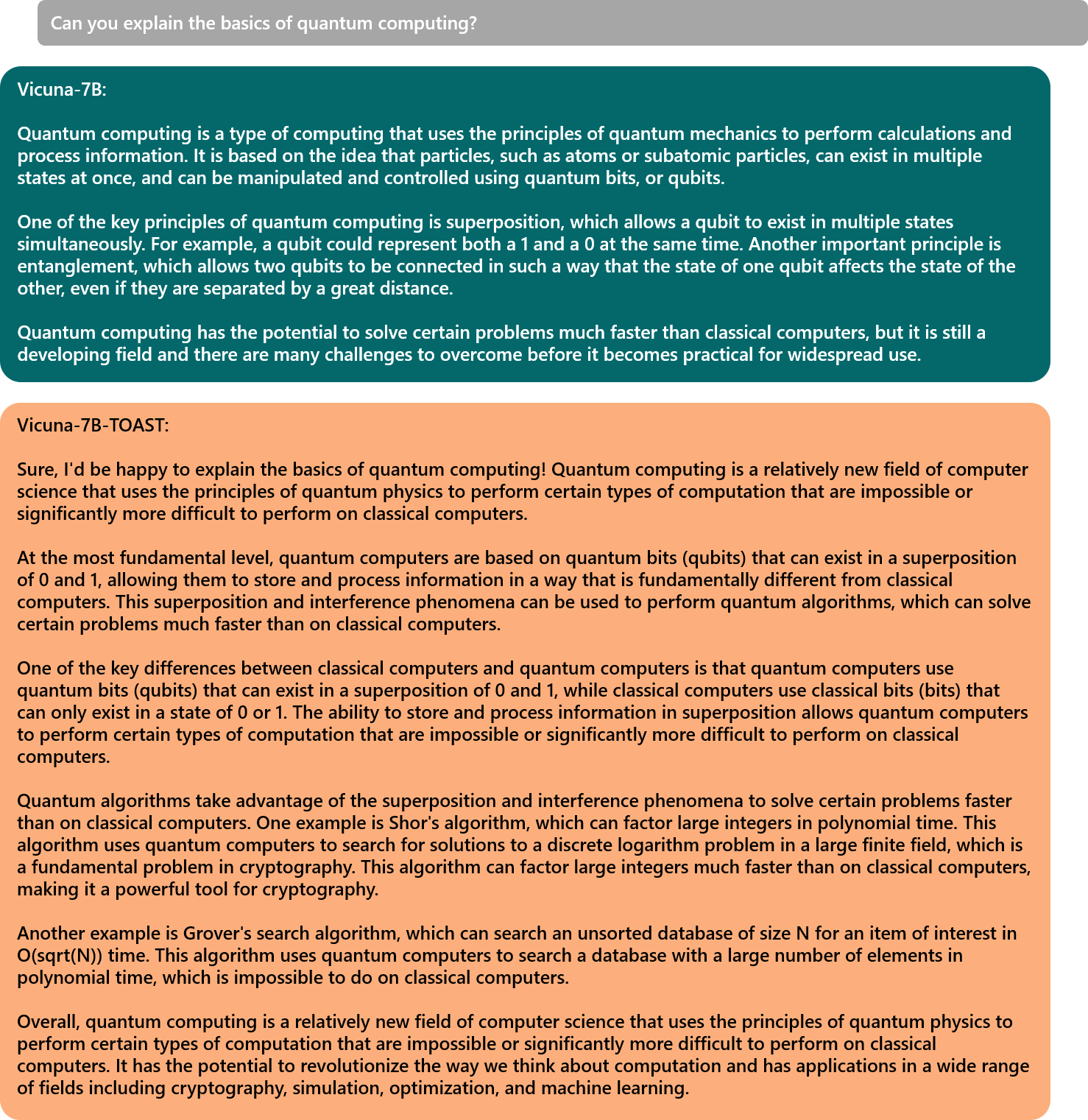}
  \caption{The responses given the question \emph{Can you explain the basics of quantum computing}. %
  }
  \label{fig:appendix_llama2}
  \vspace{-1em}
\end{figure}

\begin{figure}[t]
  \centering
  \includegraphics[width=0.95\linewidth]{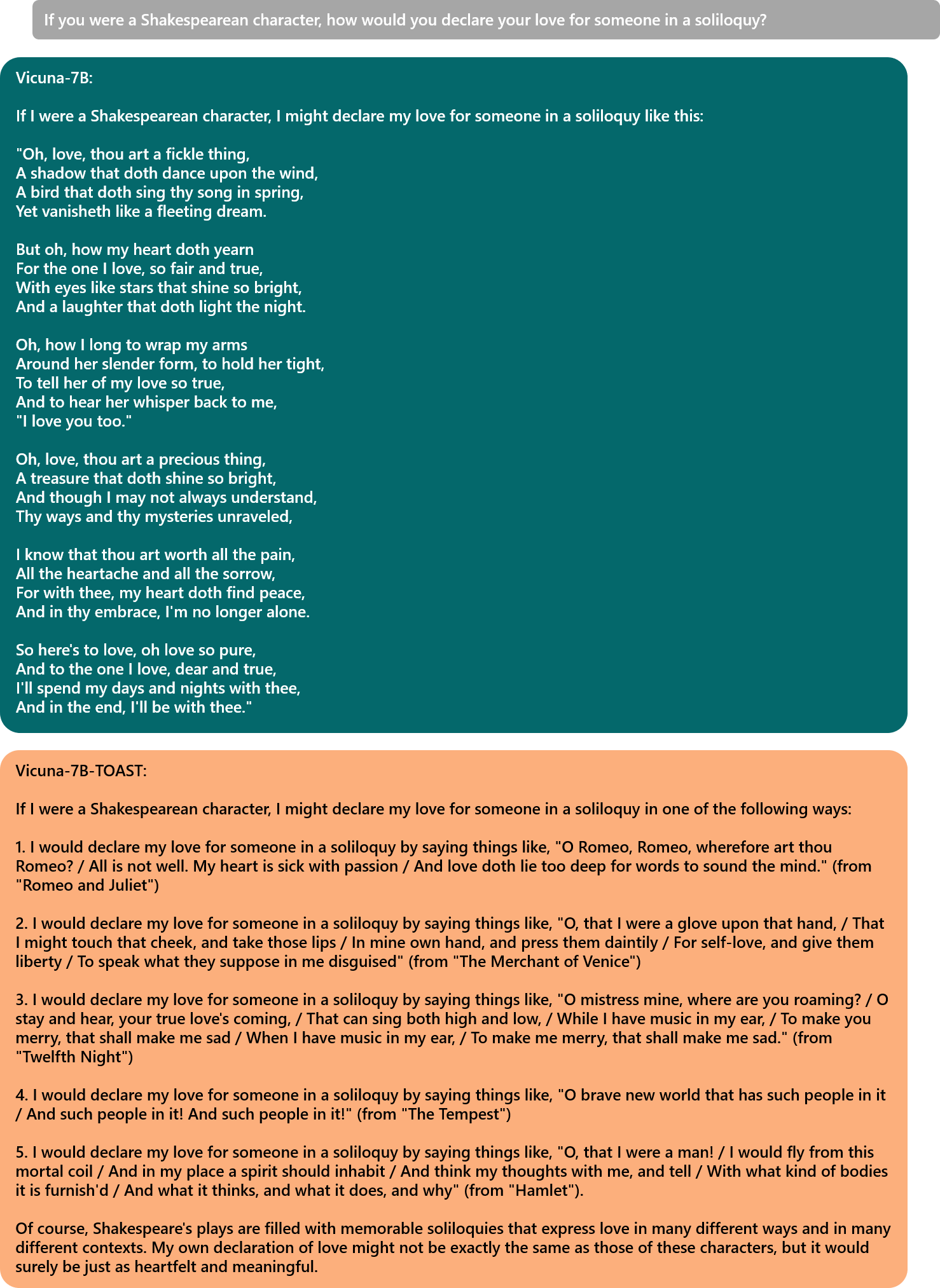}
  \caption{The responses given the question \emph{If you were a Shakespearean character, how would you declare your love for someone in a soliloquy}. %
  }
  \label{fig:appendix_llama3}
  \vspace{-1em}
\end{figure}

\begin{figure}[t]
  \centering
  \includegraphics[width=0.95\linewidth]{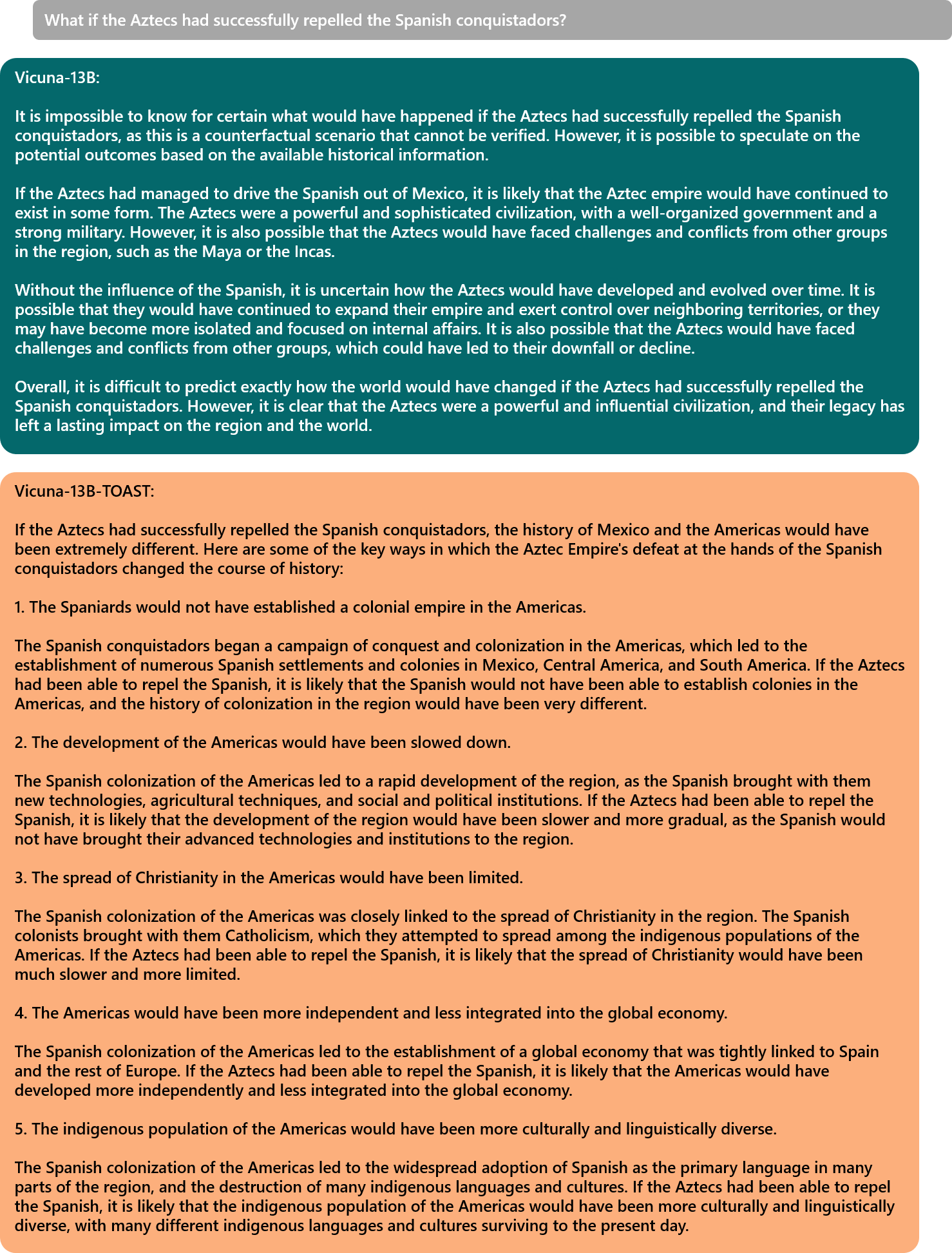}
  \caption{The responses given the question \emph{What if the Aztecs had successfully repelled the Spanish conquistadors}. %
  }
  \label{fig:appendix_llama4}
  \vspace{-1em}
\end{figure}

\begin{figure}[t]
  \centering
  \includegraphics[width=0.95\linewidth]{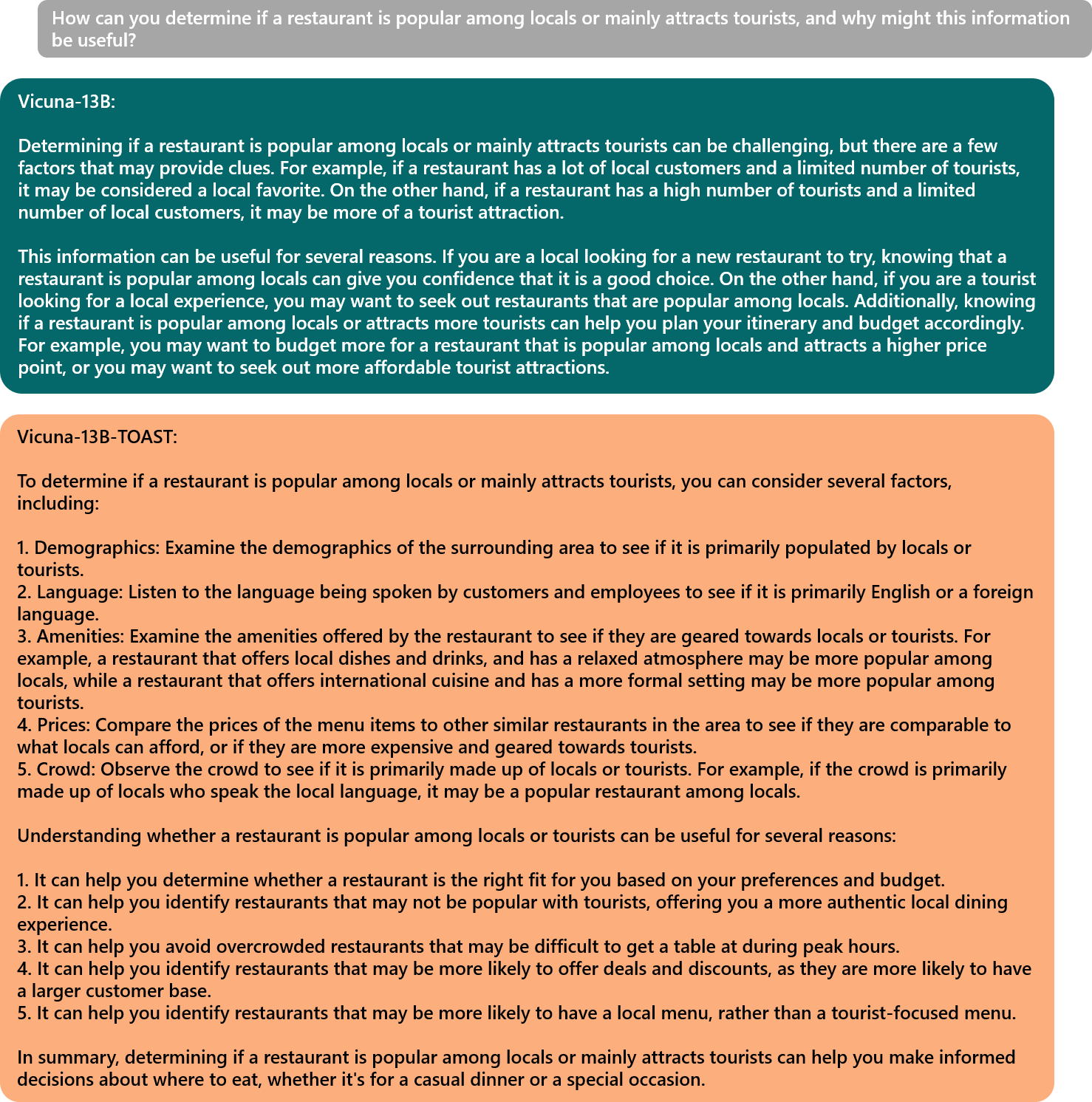}
  \caption{The responses given the question \emph{How can you determine if a restaurant is popular among locals or mainly attracts tourists, and why might this information be useful}. %
  }
  \label{fig:appendix_llama5}
  \vspace{-1em}
\end{figure}

\end{document}